\documentclass[11pt,a4paper]{article}
\usepackage[hyperref]{acl2021}
\usepackage{times}
\usepackage{latexsym}

\usepackage{microtype}

\usepackage{makecell}
\usepackage{multirow}
\usepackage{amsmath}
\usepackage{amssymb}
\usepackage{enumitem}
\usepackage{color}
\usepackage{graphicx}
\usepackage{comment}

\aclfinalcopy % Uncomment this line for the final submission
\def\aclpaperid{3936} %  Enter the acl Paper ID here

\title{Tail-to-Tail Non-Autoregressive Sequence Prediction for Chinese Grammatical Error Correction}

\author{Piji Li \ \ \ Shuming Shi \\
Tencent AI Lab, Shenzhen, China \\
\texttt{\{pijili,shumingshi\}@tencent.com}}

\date{}

\begin{document}
\maketitle
\begin{abstract}
We investigate the problem of Chinese Grammatical Error Correction (CGEC) and present a new framework named Tail-to-Tail (\textbf{TtT}) non-autoregressive sequence prediction to address the deep issues hidden in CGEC. Considering that most tokens are correct and can be conveyed directly from source to target, and the error positions can be estimated and corrected based on the bidirectional context information, thus we employ a BERT-initialized Transformer Encoder as the backbone model to conduct information modeling and conveying. Considering that only relying on the same position substitution cannot handle the variable-length correction cases, various operations such substitution, deletion, insertion, and local paraphrasing are required jointly. Therefore, a Conditional Random Fields (CRF) layer is stacked on the up tail to conduct non-autoregressive sequence prediction by modeling the token dependencies. Since most tokens are correct and easily to be predicted/conveyed to the target, then the models may suffer from a severe class imbalance issue. To alleviate this problem, focal loss penalty strategies are integrated into the loss functions. Moreover, besides the typical fix-length error correction datasets, we also construct a variable-length corpus to conduct experiments. Experimental results on standard datasets, especially on the variable-length datasets, demonstrate the effectiveness of TtT in terms of sentence-level Accuracy, Precision, Recall, and F1-Measure on tasks of error Detection and Correction\footnote{Code: \url{https://github.com/lipiji/TtT}}.

%(2) Homophonous character confusion is the fundamental reason that causes the issue of spelling error and can be corrected by substitution operations without changing the whole sequence structure. (3) Sometimes we need to delete, insert, or event paraphrase a local short subsequence to fix the grammatical errors where the sequence structure need to be adjusted.
%Although many approaches have been proposed during the past few years, some of the characteristics mentioned above are ignored and not well investigated. In this paper, we propose a new paradigm framework named tail-to-tail non-autoregressive sequence prediction (\textbf{TtT}) for the problem of CGC.
%Specifically, a BERT based sequence encoder is introduced to conduct the bidirectional representation learning for the input tokens. Then a Conditional Random Fields (CRF) layer is stacked on the tail to conduct the non-autoregressive sequence prediction by modeling the dependencies. 
%Low-rank decomposition and beamed Viterbi algorithm are introduced to accelerate the computations. 
%Focal loss penalty strategy is adopted to alleviate the class imbalance problem considering that most of the tokens in a sentence are not changed. Experimental results on several data sets, especially on the variable-length grammatical correction datasets, demonstrate the effectiveness of the proposed approach.

\end{abstract}

\section{Introduction}
\label{sec:intro}

Grammatical Error Correction (GEC) aims to automatically detect and correct the grammatical errors that can be found in a sentence \cite{DBLP:journals/corr/abs-2005-06600}. It is a crucial and essential application task in many natural language processing scenarios such as writing assistant~\cite{ghufron2018role,DBLP:conf/eacl/TetreaultSN17,DBLP:conf/bea/OmelianchukACS20}, search engine~\cite{DBLP:conf/tal/MartinsS04,DBLP:conf/coling/GaoLMQS10,DBLP:conf/www/DuanH11}, speech recognition systems~\cite{DBLP:conf/chi/KaratHHK99,DBLP:conf/interspeech/WangDLLAL20,DBLP:journals/corr/abs-2001-03041}, etc. Grammatical errors may appear in all languages~\cite{DBLP:conf/bea/DaleAN12,DBLP:conf/conll/XingWWCZ13,DBLP:conf/conll/NgWBHSB14,DBLP:conf/wanlp/RozovskayaBHZOM15,DBLP:conf/bea/BryantFAB19}, in this paper, we only focus to tackle the problem of \textit{Chinese Grammatical Error Correction (CGEC)} \cite{chang1995new}.

\begin{figure}[t!]
\centering
\includegraphics[width=\columnwidth,height=0.35\columnwidth]{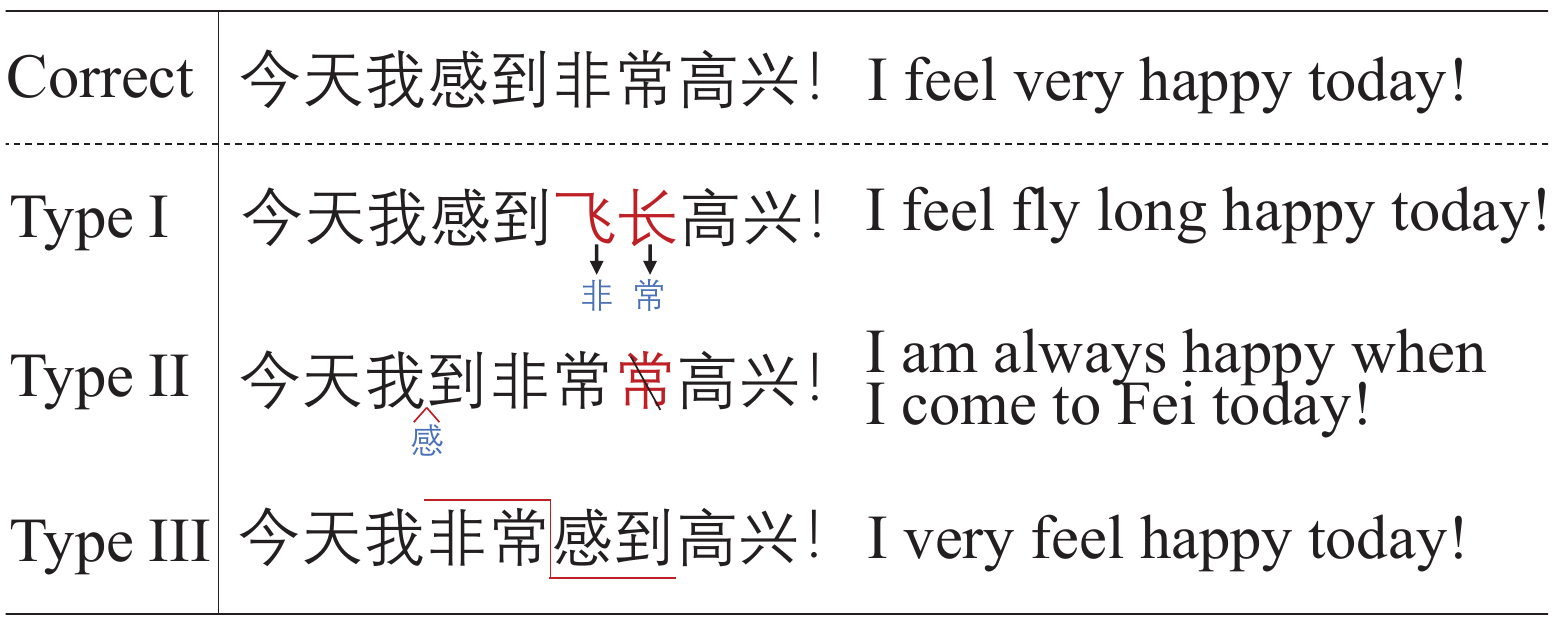}
\caption{Illustration for the three types of operations to correct the grammatical errors: Type I-substitution; Type II-deletion and insertion; Type III-local paraphrasing.}
\label{fig:error_types}
\end{figure}

\begin{figure*}[t!]
\centering
\includegraphics[width=1.7\columnwidth,height=0.5\columnwidth]{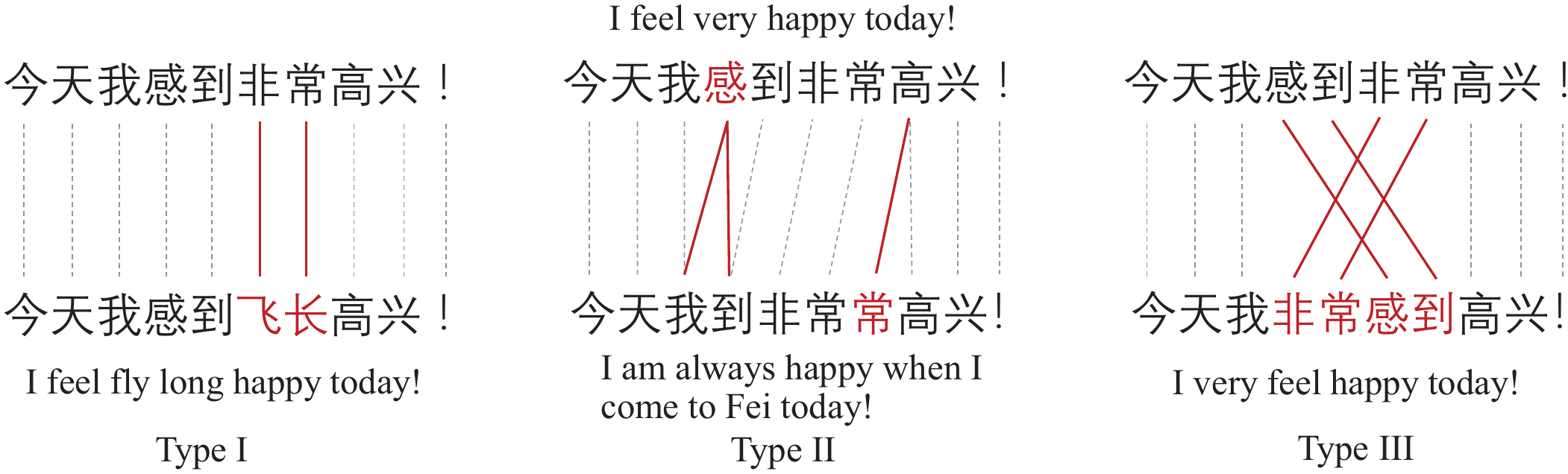}
\caption{Illustration of the token information flows from the bottom tail to the up tail. }
\label{fig:tail2tail}
\end{figure*}

We investigate the problem of CGEC and the related corpora from SIGHAN \cite{DBLP:conf/acl-sighan/TsengLCC15} and NLPCC \cite{DBLP:conf/nlpcc/ZhaoJS018} carefully, and we conclude that the grammatical error types as well as the corresponding correction operations can be categorised into three folds, as shown in Figure~\ref{fig:error_types}: (1) \textbf{Substitution}. In reality, Pinyin is the most popular input method used for Chinese writings. Thus, the homophonous character confusion (For example, in the case of Type I, the pronunciation of the wrong and correct words are both ``FeiChang'') is the fundamental reason which causes grammatical errors (or spelling errors) and can be corrected by substitution operations without changing the whole sequence structure (e.g., length). Thus, substitution is a \textit{fixed-length (FixLen) operation}. (2) \textbf{Deletion} and \textbf{Insertion}. These two operations are used to handle the cases of word redundancies and omissions respectively. (3) \textbf{Local paraphrasing}. Sometimes, light operations such as substitution, deletion, and insertion cannot correct the errors directly, therefore, a slightly subsequence paraphrasing is required to reorder partial words of the sentence, the case is shown in Type III of Figure~\ref{fig:error_types}. Deletion, insertion, and local paraphrasing can be regarded as \textit{variable-length (VarLen) operations} because they may change the sentence length. 

However, over the past few years, although a number of methods have been developed to deal with the problem of CGEC, some crucial and essential aspects are still uncovered. Generally, sequence translation and sequence tagging are the two most typical technical paradigms to tackle the problem of CGEC. Benefiting from the development of neural machine translation~\cite{DBLP:journals/corr/BahdanauCB14,DBLP:conf/nips/VaswaniSPUJGKP17}, attention-based seq2seq encoder-decoder frameworks have been introduced to address the CGEC problem in a sequence translation manner~\cite{DBLP:conf/emnlp/WangSLHZ18,DBLP:journals/corr/abs-1807-01270,DBLP:conf/acl/WangTZ19,DBLP:conf/ijcnlp/WangKKK20,DBLP:conf/acl/KanekoMKSI20}. Seq2seq based translation models are easily to be trained and can handle all types of correcting operations above mentioned. However, considering the exposure bias issue \cite{DBLP:journals/corr/RanzatoCAZ15,DBLP:conf/acl/ZhangFMYL19}, the generated results usually suffer from the phenomenon of hallucination \cite{DBLP:conf/acl/NieYWPL19,DBLP:conf/acl/MaynezNBM20} and cannot be faithful to the source text, even though copy mechanisms~\cite{DBLP:conf/acl/GuLLL16} are incorporated~\cite{DBLP:conf/acl/WangTZ19}.
Therefore, \citet{DBLP:conf/bea/OmelianchukACS20} and \citet{liang-etal-2020-bert} propose to purely employ tagging to conduct the problem of GEC instead of generation. All correcting operations such as deletion, insertion, and substitution can be guided by the predicted tags. Nevertheless, the pure tagging strategy requires to extend the vocabulary $\mathcal{V}$ to about three times by adding ``insertion-'' and ``substitution-'' prefixes to the original tokens (e.g., ``insertion-good'', ``substitution-paper'') which decrease the computing efficiency dramatically. Moreover, the pure tagging framework needs to conduct multi-pass 
prediction until no more operations are predicted, which is inefficient and less elegant. Recently, many researchers fine-tune the pre-trained language models such as BERT on the task of CGEC and obtain reasonable results~\cite{DBLP:conf/naacl/ZhaoWSJL19,DBLP:conf/aclnut/HongYHLL19,DBLP:conf/acl/ZhangHLL20}. However, limited by the BERT framework, most of them can only address the fixed-length correcting scenarios and cannot conduct deletion, insertion, and local paraphrasing operations flexibly.

Moreover, during the investigations, we also observe an obvious but crucial phenomenon for CGEC that most words in a sentence are correct and need not to be changed. This phenomenon is depicted in Figure~\ref{fig:tail2tail}, where the operation flow is from the bottom tail to the up tail. Grey dash lines represent the ``Keep'' operations and the red solid lines indicate those three types of correcting operations mentioned above. On one side, intuitively, the target CGEC model should have the ability of directly moving the correct tokens from bottom tail to up tail, then Transformer\cite{DBLP:conf/nips/VaswaniSPUJGKP17} based encoder (say BERT) seems to be a preference. On the other side, considering that almost all typical CGEC models are 
built based on the paradigms of sequence tagging or sequence translation, Maximum Likelihood Estimation (MLE)~\cite{myung2003tutorial} is usually used as the parameter learning approach, which in the scenario of CGEC, will suffer from a severe class/tag imbalance issue. However, no previous works investigate this problem thoroughly on the task of CGEC.

\begin{figure*}[t!]
\centering
\includegraphics[width=1.7\columnwidth]{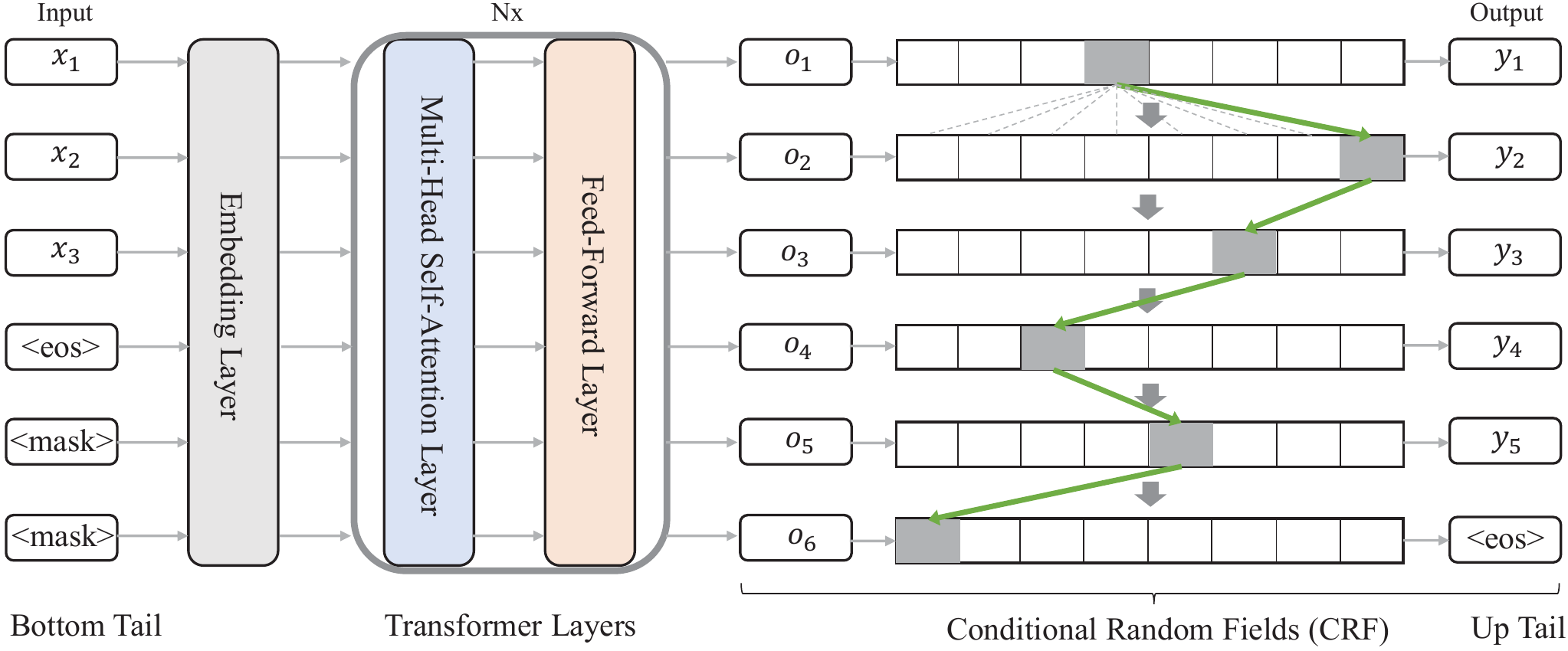}
\caption{The proposed tail-to-tail non-autoregressive sequence prediction framework (TtT).}
\label{fig:framework}
\end{figure*}

To conquer all above-mentioned challenges, we propose a new framework named tail-to-tail non-autoregressive sequence prediction, which abbreviated as \textbf{TtT}, for the problem of CGEC.
Specifically, to directly move the token information from the bottom tail to the up tail, a BERT based sequence encoder is introduced to conduct bidirectional representation learning. In order to conduct substitution, deletion, insertion, and local paraphrasing simultaneously, inspired by \cite{DBLP:conf/nips/SunLWHLD19,DBLP:conf/eacl/SuCWVBLC21}, a Conditional Random Fields (CRF) \cite{DBLP:conf/icml/LaffertyMP01} layer is stacked on the up tail to conduct non-autoregressive sequence prediction by modeling the dependencies among neighbour tokens.
%Low-rank decomposition and beamed Viterbi algorithm are introduced to accelerate the computations.
Focal loss penalty strategy~\cite{DBLP:journals/pami/LinGGHD20} is adopted to alleviate the class imbalance problem considering that most of the tokens in a sentence are not changed.
%Extensive experiments on several benchmark datasets, especially on the variable-length grammatical correction datasets, demonstrate the effectiveness of the proposed approach.
In summary, our contributions are as follows:
\begin{itemize}[topsep=0pt]
    \setlength\itemsep{-0.5em}
    \item A new framework named tail-to-tail non-autoregressive sequence prediction (TtT) is proposed to tackle the problem of CGEC.
    \item BERT encoder with a CRF layer is employed as the backbone, which can conduct substitution, deletion, insertion, and local paraphrasing simultaneously.
    \item Focal loss penalty strategy is adopted to alleviate the class imbalance problem considering that most of the tokens in a sentence are not changed.
    \item Extensive experiments on several benchmark datasets, especially on the variable-length grammatical correction datasets, demonstrate the effectiveness of the proposed approach.
\end{itemize}

\section{The Proposed TtT Framework}
\label{sec:ttt}

\subsection{Overview}
Figure~\ref{fig:framework} depicts the basic components of our proposed framework TtT. Input is an incorrect sentence $X = (x_1, x_2, \dots, x_T)$ which contains grammatical errors, where $x_i$ denotes each token (Chinese character) in the sentence, and $T$ is the length of $X$. The objective of the task grammatical error correction is to correct all errors in $X$ and generate a new sentence $Y = (y_1, y_2, \dots, y_{T'})$. Here, it is important to emphasize that $T$ is not necessary equal to $T'$. Therefore, $T'$ can be $=$, $>$, or $<$ $T$. Bidirectional semantic modeling and bottom-to-up directly token information conveying are conducted by several Transformer~\cite{DBLP:conf/nips/VaswaniSPUJGKP17} layers. A Conditional Random Fields (CRF) \cite{DBLP:conf/icml/LaffertyMP01} layer is stacked on the up tail to conduct the non-autoregressive sequence generation by modeling the dependencies among neighboring tokens.
Low-rank decomposition and beamed Viterbi algorithm are introduced to accelerate the computations.
Focal loss penalty strategy~\cite{DBLP:journals/pami/LinGGHD20} is adopted to alleviate the class imbalance problem during the training stage.

\subsection{Variable-Length Input}
Since the length $T'$ of the target sentence $Y$ is not necessary equal to the length $T$ of the input sequence $X$. Then in the training and inference stage, different length will affect the completeness of the predicted sentence, especially when $T < T'$. To handle this issue, several simple tricks are designed to pre-process the samples. Assuming $X = (x_1, x_2, x_3, \texttt{<eos>})$: (1) When $T = T'$, i.e., $Y = (y_1, y_2, y_3, \texttt{<eos>})$, then do nothing; (2) When $T > T'$, say $Y = (y_1, y_2, \texttt{<eos>})$, which means that some tokens in $X$ will be deleted during correcting. Then in the training stage, we can pad $T - T'$ special tokens $\texttt{<pad>}$ to the tail of $Y$ to make $T = T'$, then
\[
   Y = (y_1, y_2, \texttt{<eos>}, \texttt{<pad>});
\]
(3) When $T < T'$, say
\[
   Y = (y_1, y_2, y_3, y_4, y_5, \texttt{<eos>}),
\]
which means that more information should be inserted into the original sentence $X$. Then, we will pad the special symbol $\texttt{<mask>}$ to the tail of $X$ to indicate that these positions possibly can be translated into some new real tokens:
\[
   X = (x_1, x_2, x_3, \texttt{<eos>}, \texttt{<mask>}, \texttt{<mask>}).
\]

\subsection{Bidirectional Semantic Modeling}
Transformer layers \cite{DBLP:conf/nips/VaswaniSPUJGKP17} are particularly well suited to be employed to conduct the bidirectional semantic modeling and bottom-to-up information conveying. As shown in Figure~\ref{fig:framework}, after preparing the input samples, an embedding layer and a stack of Transformer layers initialized with a pre-trained Chinese BERT \cite{DBLP:conf/naacl/DevlinCLT19} are followed to conduct the semantic modeling.

Specifically, for the input, we first obtain the representations by summing the word embeddings with the positional embeddings:
\begin{equation}
    \mathbf{H}^0_t = \mathbf{E}_{w_t}+\mathbf{E}_{p_t}
\end{equation}
where $0$ is the layer index and  $t$ is the state index. $\mathbf{E}_w$ and $\mathbf{E}_p$ are the embedding vectors for tokens and positions, respectively.

Then the obtained embedding vectors $\mathbf{H}^0$ are fed into several Transformer layers. Multi-head self-attention is used to conduct bidirectional representation learning:
\begin{equation}
\begin{split}
\mathrm{\bf H}^{1}_{t} &= \textsc{Ln}\left(\textsc{Ffn} (\mathrm{\bf H}^{1}_{t}) +\mathrm{\bf H}^1_t \right) \\
\mathrm{\bf H}^{1}_{t} &= \textsc{Ln}\left(\textsc{Slf-Att} (\mathrm{\bf Q}^{0}_{t}, \mathrm{\bf K}^{0}, \mathrm{\bf V}^{0}) +\mathrm{\bf H}^0_t \right) \\
\mathrm{\bf Q}^{0} &=  \mathrm{\bf H}^{0} \mathrm{\bf W}^{Q} \\
\mathrm{\bf K}^{0}, \mathrm{\bf V}^{0} &= \mathrm{\bf H}^{0}\mathrm{\bf W}^{K}, \mathrm{\bf H}^{0} \mathrm{\bf W}^{V}
\end{split}
\label{eql:formant_c1}
\end{equation}
where \textsc{Slf-Att}($\cdot$), \textsc{Ln}($\cdot$), and \textsc{Ffn}($\cdot$) represent self-attention mechanism, layer normalization, and feed-forward network respectively \cite{DBLP:conf/nips/VaswaniSPUJGKP17}. Note that our model is a non-autoregressive sequence prediction framework, thus we use all the sequence states $\mathrm{\bf K}^{0}$ and $\mathrm{\bf V}^{0}$ as the attention context. Then each node will absorb the context information bidirectionally.
After $L$ Transformer layers, we obtain the final output representation vectors $\mathrm{\bf H}^{L} \in \mathbb{R}^{max(T,T') \times d}$. 

\subsection{Non-Autoregressive Sequence Prediction}

\paragraph{Direct Prediction}

The objective of our model is to translate the input sentence $X$ which contains grammatical errors into a correct sentence $Y$. Then, since we have obtained the sequence representation vectors $\mathrm{\bf H}^{L}$, we can directly add a \texttt{softmax} layer to predict the results, just similar to the methods used in non-autoregressive neural machine translation~\cite{DBLP:journals/corr/abs-2012-15833} and BERT-based fine-tuning framework for the task of grammatical error correction~\cite{DBLP:conf/naacl/ZhaoWSJL19,DBLP:conf/aclnut/HongYHLL19,DBLP:conf/acl/ZhangHLL20}.

Specifically, a linear transformation layer is plugged in and \texttt{softmax} operation is utilized to generate a probability distribution $P_\mathrm{dp}(y_t)$ over the target vocabulary $\mathcal{V}$:
\begin{align}
\begin{split}
    \mathbf{s}_t &= \mathbf{h}^\top_t\mathbf{W}_s + \mathbf{b}_s\\
    P_\mathrm{dp}(y_t) &= \texttt{softmax}(\mathbf{s}_t)
\end{split}
\label{eq:s}
\end{align}
where $\mathbf{h_t} \in \mathbb{R}^d$,  $\mathbf{W}_s \in \mathbb{R}^{d\times|\mathcal{V}|}$,  $\mathbf{b}_s \in \mathbb{R}^{|\mathcal{V}|}$, and $\mathbf{s}_t \in \mathbb{R}^{|\mathcal{V}|}$. 
Then we obtain the result for each state based on the predicted distribution:
\begin{equation}
y'_t = \texttt{argmax}(P_\mathrm{dp}(y_t))
\end{equation}
However, although this direct prediction method is effective on the fixed-length grammatical error correction problem, it can only conduct the same-positional substitution operation. For complex correcting cases which require deletion, insertion, and local paraphrasing, the performance is unacceptable. This inferior performance phenomenon is also discussed in the tasks of non-autoregressive neural machine translation~\cite{DBLP:journals/corr/abs-2012-15833}.

One of the essential reasons causing the inferior performance is that the dependency information among the neighbour tokens are missed. Therefore, dependency modeling should be called back to improve the performance of generation. Naturally, linear-chain CRF \cite{DBLP:conf/icml/LaffertyMP01} is introduced to fix this issue, and luckily, \citet{DBLP:conf/nips/SunLWHLD19,DBLP:conf/eacl/SuCWVBLC21} also employ CRF to address the problem of non-autoregressive sequence generation, which inspired us a lot. 

\paragraph{Dependency Modeling via CRF}
Then given the input sequence $X$, under the CRF framework, the likelihood of the target sequence $Y$ with length $T'$ is constructed as:
\begin{align}
\begin{split}
    P_\mathrm{crf}&({Y}|{X}) = \\
    &\frac{1}{Z({X})}\exp\left(\sum_{t=1}^{T^{\prime}}s(y_t) + \sum_{t=2}^{T^{\prime}}t(y_{t-1},y_t)\right)
\end{split}
\end{align}
where $Z(X)$ is the normalizing factor and $s(y_t)$ represents the label score of $y$ at position $t$, which can be obtained from the predicted logit vector $\mathbf{s}_t \in \mathbb{R}^{|\mathcal{V}|}$ from Eq. (\ref{eq:s}), i.e., $\mathbf{s}_t(\mathcal{V}^{y_t})$, where $\mathcal{V}^{y_t}$ is the vocabulary index of token $y_t$. The value $t(y_{t-1},y_t)={\bf M}_{y_{t-1},y_t}$ denotes the transition score from token $y_{t-1}$ to $y_t$ where ${\bf M}\in\mathbb{R}^{|\mathcal{V}|\times |\mathcal{V}|}$ is the transition matrix, which is the core term to conduct dependency modeling. Usually, $\mathbf{M}$ can be learnt as neural network parameters during the end-to-end training procedure. However, $|\mathcal{V}|$ is typically very large especially in the text generation scenarios (more than $32k$), therefore it is infeasible to obtain $\mathbf{M}$ and $Z(X)$ efficiently in practice.
To overcome this obstacle, as the method used in \cite{DBLP:conf/nips/SunLWHLD19,DBLP:conf/eacl/SuCWVBLC21}, we introduce two low-rank neural parameter metrics $\mathbf{E}_1$, $\mathbf{E}_2 \in \mathbb{R}^{|\mathcal{V}|\times d_m}$ to approximate the full-rank transition matrix $\mathbf{M}$ by:
\begin{equation}
\mathbf{M} = \mathbf{E}_1\mathbf{E}_2^\top
\end{equation}
where $d_m \ll |\mathcal{V}|$.
To compute the normalizing factor $Z(X)$, the original Viterbi algorithm \cite{forney1973viterbi,DBLP:conf/icml/LaffertyMP01} need to search all paths. To improve the efficiency, here we only visit the truncated top-$k$ nodes at each time step approximately \cite{DBLP:conf/nips/SunLWHLD19,DBLP:conf/eacl/SuCWVBLC21}.

\subsection{Training with Focal Penalty}

Considering the characteristic of the directly bottom-to-up information conveying of the task CGEC, therefore, both tasks, direct prediction and CRF-based dependency modeling, can be incorporated jointly into a unified framework during the training stage. The reasons are that, intuitively, direct prediction will focus on the fine-grained predictions at each position, while CRF-layer will pay more attention to the high-level quality of the whole global sequence.
We employ Maximum Likelihood Estimation (MLE) to conduct parameter learning and treat negative log-likelihood (NLL) as the loss function. Thus, the optimization objective for direct prediction $\mathcal{L}_{\mathrm{dp}}$ is:
\begin{equation}
\begin{split}
\mathcal{L}_{\mathrm{dp}} = -\sum^{T'}_{t=1} \log P_{\mathrm{dp}}(\mathrm{y}_t|X)
\end{split}
\label{eq:nll_dp}
\end{equation}
And the loss function $\mathcal{L}_{\mathrm{crf}}$ for CRF-based dependency modeling is:
\begin{equation}
\begin{split}
\mathcal{L}_{\mathrm{crf}} = -\log P_\mathrm{crf}({Y}|{X})
\end{split}
\label{eq:nll_crf}
\end{equation}
Then the final optimization objective is:
\begin{equation}
\mathcal{L} = \mathcal{L}_{\mathrm{dp}} + \mathcal{L}_{\mathrm{crf}}
\end{equation}

As mentioned in Section~\ref{sec:intro}, one obvious but crucial phenomenon for CGEC is that most words in a sentence are correct and need not to be changed. Considering that maximum likelihood estimation is used as the parameter learning approach in those two tasks, then a simple copy strategy can lead to a sharp decline in terms of loss functions. Then, intuitively, the grammatical error tokens which need to be correctly fixed in practice, unfortunately, attract less attention during the training procedure. Actually, these tokens, instead, should be regarded as the focal points and contribute more to the optimization objectives. However, no previous works investigate this problem thoroughly on the task of CGEC.

To alleviate this issue, we introduce a useful trick, focal loss~\cite{DBLP:journals/pami/LinGGHD20} , into our loss functions for direct prediction and CRF:
\begin{align}
\begin{split}
\mathcal{L}_{\mathrm{dp}}^{\mathrm{fl}} &= -\sum^{T'}_{t=1} (1 - P_{\mathrm{dp}}(\mathrm{y}_t|X))^{\gamma} \log P_{\mathrm{dp}}(\mathrm{y}_t|X)\\
\mathcal{L}_{\mathrm{crf}}^{\mathrm{fl}} &= -(1 - P_\mathrm{crf}({Y}|{X}))^{\gamma}\log P_\mathrm{crf}({Y}|{X})
\end{split}
\end{align}
where $\gamma$ is a hyperparameter to control the penalty weight. It is obvious that $\mathcal{L}_{\mathrm{dp}}^{\mathrm{fl}}$ is penalized on the token level, while $\mathcal{L}_{\mathrm{crf}}^{\mathrm{fl}}$ is weighted on the sample level and will work in the  condition of batch-training. The final optimization objective with focal penalty strategy is:
\begin{equation}
\mathcal{L}^{\mathrm{fl}} = \mathcal{L}_{\mathrm{dp}}^{\mathrm{fl}} + \mathcal{L}_{\mathrm{crf}}^{\mathrm{fl}}
\end{equation}

\subsection{Inference}
During the inference stage, for the input source sentence $X$, we can employ the original $|\mathcal{V}|$ nodes Viterbi algorithm to obtain the target global optimal result. We can also utilize the truncated top-$k$ Viterbi algorithm for high computing efficiency \cite{DBLP:conf/nips/SunLWHLD19,DBLP:conf/eacl/SuCWVBLC21}.

\section{Experimental Setup}
\subsection{Settings}
The core technical components of our proposed TtT is Transformer~\cite{DBLP:conf/nips/VaswaniSPUJGKP17} and CRF~\cite{DBLP:conf/icml/LaffertyMP01}. 
The pre-trained Chinese BERT-base model \cite{DBLP:conf/naacl/DevlinCLT19} is employed to initialize the model.
To approximate the transition matrix in the CRF layer, we set the dimension $d$ of matrices ${\bf E}_1$ and ${\bf E}_2$ as 32. For the normalizing factor ${\bf Z}({\bf X})$, we set the predefined beam size $k$ as 64.
The hyperparameter $\gamma$ which is used to weight the focal penalty term is set to $0.5$ after parameter tuning.
Training batch-size is 100, learning rate is $1e-5$, dropout rate is $0.1$. 
Adam optimizer \cite{DBLP:journals/corr/KingmaB14} is used to conduct the parameter learning.

\subsection{Datasets}

\begin{table}[!t]
\centering
\small
\begin{tabular}{l|c|c|c|c}
\Xhline{2\arrayrulewidth}
 Corpus & \#Train & \#Dev & \#Test & Type \\
 \hline
 SIGHAN15 & 2,339 &  - &  1,100 & FixLen  \\
 \hline
 HybirdSet & 274,039 & 3,162 & 3,162 & FixLen    \\
 \hline
 TtTSet & 539,268 & 5,662 & 5,662 & VarLen \\
 \hline
 \Xhline{2\arrayrulewidth}
\end{tabular}
\caption{Statistics of the datasets.}
\label{tab:datasets}
\end{table}

The overall statistic information of the datasets used in our experiments are depicted in Table~\ref{tab:datasets}.

\noindent \textbf{SIGHAN15~\cite{DBLP:conf/acl-sighan/TsengLCC15}\footnote{\url{http://ir.itc.ntnu.edu.tw/lre/sighan8csc.html}}} This is a benchmark dataset for the evaluation of CGEC and it contains 2,339 samples for training and 1,100 samples for testing. As did in some typical previous works~\cite{DBLP:conf/acl/WangTZ19,DBLP:conf/acl/ZhangHLL20}, we also use the SIGHAN15 testset as the benchmark dataset to evaluate the performance of our models as well as the baseline methods in fixed-length (FixLen) error correction settings. 

\noindent\textbf{HybirdSet~\cite{DBLP:conf/emnlp/WangSLHZ18}\footnote{\url{https://github.com/wdimmy/Automatic-Corpus-Generation}}} It is a newly released dataset constructed according to a prepared confusion set based on the results of ASR \cite{yu2014automatic} and OCR~\cite{DBLP:conf/acl-vlc/TongE96}. This dataset contains about 270k paired samples and it is also a FixLen dataset.

\noindent\textbf{TtTSet} Considering that datasets of \texttt{SIGHAN15} and \texttt{HybirdSet} are all FixLen type datasets, in order to demonstrate the capability of our model TiT on the scenario of Variable-Length (VarLen) CGEC, based on the corpus of \texttt{HybirdSet}, we build a new VarLen dataset. Specifically, operations of deletion, insertion, and local shuffling are conducted on the original sentences to obtain the incorrect samples. Each operation covers one-third of samples, thus we get about 540k samples finally.

\begin{table*}[!t]
\small
    \centering
    \resizebox{1.8\columnwidth}{!}{
    \begin{tabular}{l|cccc|cccc}
    \Xhline{3\arrayrulewidth}
     \multirow{2}{*}{\textbf{Model}} & \multicolumn{4}{c|}{\textbf{Detection}} & \multicolumn{4}{c}{\textbf{Correction}} \\ 
     \cline{2-9} & \textsc{Acc.} & \textsc{Prec.} & \textsc{Rec.} & \textsc{F1} & \textsc{Acc.} & \textsc{Prec.} & \textsc{Rec.} & \textsc{F1} \\
     \hline
    NTOU (\citeyear{DBLP:conf/acl-sighan/TsengLCC15}) & 42.2 & 42.2 & 41.8 & 42.0 & 39.0 & 38.1 & 35.2 & 36.6 \\
    NCTU-NTUT (\citeyear{DBLP:conf/acl-sighan/TsengLCC15}) & 60.1 & 71.7 & 33.6 & 45.7 & 56.4 & 66.3 & 26.1 & 37.5 \\
    HanSpeller++ (\citeyear{DBLP:conf/acl-sighan/ZhangXHZC15}) & 70.1 & 80.3 & 53.3 & 64.0 & 69.2 & 79.7 & 51.5 & 62.5 \\
    Hybird (\citeyear{DBLP:conf/emnlp/WangSLHZ18}) & - & 56.6 & 69.4 & 62.3 & - & - 
    & - & 57.1 \\
    FASPell (\citeyear{DBLP:conf/aclnut/HongYHLL19}) & 74.2 & 67.6 & 60.0 & 63.5 & 73.7 & 66.6 & 59.1 & 62.6 \\
    Confusionset (\citeyear{DBLP:conf/acl/WangTZ19}) & - & 66.8 & 73.1 & 69.8 & - & 71.5 & 59.5 & 64.9 \\
    SoftMask-BERT (\citeyear{DBLP:conf/acl/ZhangHLL20}) & 80.9 & 73.7 & 73.2 & 73.5 & 77.4 & 66.7 & 66.2 & 66.4 \\
    Chunk (\citeyear{DBLP:conf/emnlp/BaoLW20}) & 76.8 & 88.1 & 62.0 & 72.8 & 74.6 & 87.3 & 57.6 & 69.4 \\
    %SpellGCN (\citeyear{DBLP:conf/acl/ChengXCJWWCQ20}) & - & 74.8 & \textbf{80.7} & 77.7 & - & 72.1 & \textbf{77.7} & 75.9 \\
    \textbf{*SpellGCN} (\citeyear{DBLP:conf/acl/ChengXCJWWCQ20}) & - & 85.9 & 80.6 & 83.1 & - & 85.4 & 77.6 & 81.3 \\
    \hline
    %BERT-Threshold & 60.8 & 65.4 & 43.5 & 52.3 & 57.4 & 61.3 & 36.5 & 45.8 \\
    Transformer-s2s (Sec.\ref{sec:baselines}) & 67.0 & 73.1 & 52.2 & 50.9 & 66.2 & 72.5 & 50.6 & 59.6\\
    GPT2-finetune (Sec.\ref{sec:baselines}) & 65.1 & 70.0 & 51.9 & 59.4 & 64.6 & 69.1 & 50.7 & 58.5\\
    BERT-finetune (Sec.\ref{sec:baselines}) & 75.4 & 84.1 & 61.5 & 71.1 & 71.6 & 82.2 & 53.9 & 65.1\\
    \hline
    %BERT-Threshold & 58.5 & 61.7 & 41.3 & 49.5 & 53.6 & 55.2 & 31.6 & 40.1 \\
    \textbf{TtT} (Sec.\ref{sec:ttt}) & 82.7 & 85.4 & 78.1 & 81.6 & 81.5 & 85.0 & 75.6 & 80.0 \\
    \Xhline{3\arrayrulewidth}
    \end{tabular}}
    \caption{Detection and Correction results evaluated on the SIGHAN2015 testset (1100 samples). \textbf{*} is the SOTA.} 
    \label{tbl:results_sighan}
\end{table*}

\begin{table*}[!t]
\small
    \centering
    \resizebox{1.8\columnwidth}{!}{
    \begin{tabular}{l|cccc|cccc}
    \Xhline{3\arrayrulewidth}
     \multirow{2}{*}{\textbf{Model}} & \multicolumn{4}{c|}{\textbf{Detection}} & \multicolumn{4}{c}{\textbf{Correction}} \\ 
     \cline{2-9} & \textsc{Acc.} & \textsc{Prec.} & \textsc{Rec.} & \textsc{F1} & \textsc{Acc.} & \textsc{Prec.} & \textsc{Rec.} & \textsc{F1} \\
    \hline
    Transformer-s2s (Sec.\ref{sec:baselines}) & 25.6 & 65.6 & 16.1 & 25.9 & 24.6 & 63.6 & 14.8 & 24.0\\
    GPT2-finetune (Sec.\ref{sec:baselines}) & 51.3 & 85.2 & 47.9 & 61.3 & 45.1 & 82.8 & 40.2 & 54.1\\
    BERT-finetune (Sec.\ref{sec:baselines}) & 46.8 & 89.0 & 38.9 & 54.1 & 36.9 & 84.8 & 26.7 & 40.7\\
    \hline
    \textbf{TtT} (Sec.\ref{sec:ttt}) & \textbf{55.6} & \textbf{89.8} & \textbf{50.4} & \textbf{64.6} & \textbf{60.6} & \textbf{88.5} & \textbf{44.2} & \textbf{58.9} \\
    \Xhline{3\arrayrulewidth}
    \end{tabular}}
    \caption{Detection and Correction results evaluated on the TtTSet testset (5662 samples).}
    \label{tbl:results_tttset}
\end{table*}

\subsection{Comparison Methods}
\label{sec:baselines}
We compare the performance of \textbf{TtT} with several strong baseline methods on both FixLen and VarLen settings.\\
\textbf{NTOU} employs n-gram language model with a reranking strategy to conduct prediction \cite{DBLP:conf/acl-sighan/TsengLCC15}.\\
\textbf{NCTU-NTUT} also uses CRF to conduct label dependency modeling \cite{DBLP:conf/acl-sighan/TsengLCC15}. \\
\textbf{HanSpeller++} employs Hidden Markov Model with a reranking strategy to conduct the prediction \cite{DBLP:conf/acl-sighan/ZhangXHZC15}.\\
\textbf{Hybrid} utilizes LSTM-based seq2seq framework to conduct generation \cite{DBLP:conf/emnlp/WangSLHZ18} and \textbf{Confusionset} introduces a copy mechanism into seq2seq framework~\cite{DBLP:conf/acl/WangTZ19}.\\
\textbf{FASPell} incorporates BERT into the seq2seq for better performance \cite{DBLP:conf/aclnut/HongYHLL19}.\\
\textbf{SoftMask-BERT} firstly conducts error detection using a GRU-based model and then incorporating the predicted results with the BERT model using a soft-masked strategy \cite{DBLP:conf/acl/ZhangHLL20}. Note that the best results of \textbf{SoftMask-BERT} are obtained after pre-training on a large-scale dataset with 500M paired samples.\\
\textbf{SpellGCN} proposes to incorporate phonological and visual similarity knowledge into language models via a specialized graph convolutional network~\cite{DBLP:conf/acl/ChengXCJWWCQ20}.\\
\textbf{Chunk} proposes a chunk-based decoding method with global optimization to correct single character and multi-character word typos in a unified framework~\cite{DBLP:conf/emnlp/BaoLW20}.

We also implement some classical methods for comparison and ablation analysis, especially for the VarLen correction problem.
\textbf{Transformer-s2s} is the typical Transformer-based seq2seq framework for sequence prediction \cite{DBLP:conf/nips/VaswaniSPUJGKP17}.
\textbf{GPT2-finetune} is also a sequence generation framework fine-tuned based on a pre-trained Chinese GPT2 model\footnote{\url{https://github.com/lipiji/Guyu}} \cite{radford2019language,DBLP:journals/corr/abs-2003-04195}.   
\textbf{BERT-finetune} is just fine-tune the Chinese BERT model on the CGEC corpus directly.
Beam search decoding strategy is employed to conduct generation for Transformer-s2s and GPT2-finetune, and beam-size is 5. 
Note that some of the original methods above mentioned can only work in the FixLen settings, such as \textbf{SoftMask-BERT} and \textbf{BERT-finetune}.

\subsection{Evaluation Metrics}
Following the typical previous works \cite{DBLP:conf/acl/WangTZ19,DBLP:conf/aclnut/HongYHLL19,DBLP:conf/acl/ZhangHLL20}, we employ sentence-level \textbf{Accuracy}, \textbf{Precision}, \textbf{Recall}, and \textbf{F1-Measure} as the automatic metrics to evaluate the performance of all systems\footnote{\url{http://nlp.ee.ncu.edu.tw/resource/csc.html}}. We also report the detailed results for error \textbf{Detection} (all locations of incorrect characters in a given sentence should be completely identical with the gold standard) and \textbf{Correction} (all locations and corresponding corrections of incorrect characters should be completely identical with the gold standard) respectively \cite{DBLP:conf/acl-sighan/TsengLCC15}.

\section{Results and Discussions}

\subsection{Results in FixLen Scenario}

Table~\ref{tbl:results_sighan} depicts the main evaluation results of our proposed framework \textbf{TtT} as well as the comparison baseline methods. It should be emphasized that SoftMask-BERT is pre-trained on a 500M-size paired dataset. Our model TtT, as well as the baseline methods such as Transformer-s2s, GPT2-finetune, BERT-finetune, and Hybird are all trained on the 270k-size HybirdSet. Nevertheless, TtT obtains improvements on the tasks of error Detection and Correction compared to most of the strong baselines on F1 metric, which indicates the superiority of our proposed approach.

\begin{table*}[!t]
\small
    \centering
    \resizebox{1.8\columnwidth}{!}{
    \begin{tabular}{l|c|cccc|cccc}
    \Xhline{3\arrayrulewidth}
     \multirow{2}{*}{\textbf{TrainSet}} & \multirow{2}{*}{\textbf{Model}}  & \multicolumn{4}{c|}{\textbf{Detection}} & \multicolumn{4}{c}{\textbf{Correction}} \\ 
     \cline{3-10}
     & & \textsc{Acc.} & \textsc{Prec.} & \textsc{Rec.} & \textsc{F1} & \textsc{Acc.} & \textsc{Prec.} & \textsc{Rec.} & \textsc{F1} \\
    \hline
    SIGHAN15 & Transformer-s2s & 46.5 & 42.2 & 23.6 & 30.3 & 43.4 & 34.9 & 17.3 & 23.2\\
    & GPT2-finetune & 45.2 & 42.3 & 30.8 & 35.7 & 42.6 & 37.7 & 25.5 & 30.4\\
    & BERT-finetune & 35.8 & 34.1 & 32.8 & 33.4 & 31.3 & 27.1 & 23.6 & 25.3\\
    & TtT  & 51.3 & 50.6 & 38.0 & 43.4 & 45.8 & 41.9 & 26.7 & 32.7\\
    \hline
        HybirdSet & Transformer-s2s & 67.0 & 73.1 & 52.2 & 50.9 & 66.2 & 72.5 & 50.6 & 59.6\\
    & GPT2-finetune & 65.1 & 70.0 & 51.9 & 59.4 & 64.6 & 69.1 & 50.7 & 58.5\\
    & BERT-finetune & 75.4 & 84.1 & 61.5 & 71.1 & 71.6 & 82.2 & 53.9 & 65.1\\
    & TtT  & 82.7 & 85.4 & 78.1 &81.6 & 81.5 & 85.0 & 75.6 & 80.0\\
    \Xhline{3\arrayrulewidth}
    \end{tabular}}
    \caption{Performance of models trained on different datasets.}
    \label{tbl:results_trainset}
\end{table*}

\begin{table*}[!t]
\small
    \centering
    \resizebox{1.8\columnwidth}{!}{
    \begin{tabular}{l|c|cccc|cccc}
    \Xhline{3\arrayrulewidth}
     \multirow{2}{*}{\textbf{TrainSet}} & \multirow{2}{*}{\textbf{Model}}  & \multicolumn{4}{c|}{\textbf{Detection}} & \multicolumn{4}{c}{\textbf{Correction}} \\ 
     \cline{3-10}
     & & \textsc{Acc.} & \textsc{Prec.} & \textsc{Rec.} & \textsc{F1} & \textsc{Acc.} & \textsc{Prec.} & \textsc{Rec.} & \textsc{F1} \\
    \hline
    SIGHAN15 & TtT w/o $\mathcal{L}_{\mathrm{crf}}$ & 35.8 & 34.1 & 32.8 & 33.4 & 31.3 & 27.1 & 23.6 & 25.3\\
    & TtT w/o $\mathcal{L}_{\mathrm{dp}}$ & 35.5 & 32.0 & 28.0 & 29.9 & 31.2 & 24.9 & 19.3 & 21.6\\
    & TtT  & 42.6 & 39.4 & 31.5 & 35.0 & 36.7 & 28.9 & 23.6 & 26.0\\
    \hline
        HybirdSet & TtT w/o $\mathcal{L}_{\mathrm{crf}}$ & 75.4 & 84.1 & 61.5 & 71.1 & 71.6 & 82.2 & 53.9 & 65.1\\
    & TtT w/o $\mathcal{L}_{\mathrm{dp}}$ & 81.2 & 83.4 & 77.1 & 80.1 & 80.0 & 83.0 & 74.7 & 78.6\\
    & TtT  & 82.7 & 85.6 & 77.9 &81.5 & 81.1 & 85.0 & 74.7 & 79.5\\
    \Xhline{3\arrayrulewidth}
    \end{tabular}}
    \caption{Ablation analysis of $\mathcal{L}_{\mathrm{dp}}$ and $\mathcal{L}_{\mathrm{crf}}$.}
    \label{tbl:results_loss}
\end{table*}

\begin{table*}[!t]
\small
    \centering
    \resizebox{1.65\columnwidth}{!}{
    \begin{tabular}{l|c|cccc|cccc}
    \Xhline{3\arrayrulewidth}
     \multirow{2}{*}{\textbf{TrainSet}} & \multirow{2}{*}{\textbf{$\gamma$}}  & \multicolumn{4}{c|}{\textbf{Detection}} & \multicolumn{4}{c}{\textbf{Correction}} \\ 
     \cline{3-10}
     & & \textsc{Acc.} & \textsc{Prec.} & \textsc{Rec.} & \textsc{F1} & \textsc{Acc.} & \textsc{Prec.} & \textsc{Rec.} & \textsc{F1} \\
    \hline
    SIGHAN15 & 0.0 & 42.6 & 39.4 & 31.5 & 35.0 & 36.7 & 28.9 & 23.6 & 26.0\\
    & 0.1 & 48.8 & 47.0 & 35.5 & 40.3 & 43.8 & 38.7 & 25.1 & 30.4\\
    & 0.5 & 51.3 & 50.6 & 38.0 & 43.4 & 45.8 & 41.9 & 26.7 & \textbf{32.6}\\
    & 1.0 & 51.8 & 51.3 & 37.7 & 43.5 & 46.3 & 42.5 & 26.5 & \textbf{32.6}\\
    & 2.0 & 50.0 & 48.6 & 36.3 & 41.5 & 44.4 & 39.5 & 25.0 & 30.6\\
    & 5.0 & 48.9 & 47.1 & 37.2 & 47.6 & 42.8 & 37.6 & 25.1 & 30.6\\
    \hline
        HybirdSet & 0.0 & 82.7 & 85.6 & 77.9 &81.5 & 81.1 & 85.0 & 74.7 & 79.5\\
    & 0.1 & 74.6 & 73.5 & 75.4 & 74.4 & 73.2 & 72.7 & 72.6 & 72.7\\
    & 0.5 & 82.7 & 85.4 & 78.0 & 81.6 & 81.5 & 85.0 & 75.6 & \textbf{80.0}\\
    & 1.0 & 81.1 & 83.2 & 77.1 & 80.0 & 80.0 & 82.8 & 74.9 & 78.6\\
    & 2.0 & 79.2 & 80.4 & 76.2 & 78.2 & 78.2 & 80.0 & 74.1 & 76.9\\
    & 5.0 & 80.3 & 81.6 & 77.3 & 79.4 & 78.7 & 80.9 & 74.1 & 77.4\\
    \Xhline{3\arrayrulewidth}
    \end{tabular}}
    \caption{Tuning for focal loss hyperparameter $\gamma$.}
    \label{tbl:results_g}
\end{table*}

\subsection{Results in VarLen Scenario}

Benefit from the CRF-based dependency modeling component, TtT can conduct deletion, insertion, local paraphrasing operations jointly to address the Variable-Length (VarLen) error correction problem.
The experimental results are described in Table~\ref{tbl:results_tttset}. Considering that those sequence generation methods such as Transformer-s2s and GPT2-finetune can also conduct VarLen correction operation, thus we report their results as well. From the results, we can observe that \textbf{TtT} can also achieve a superior performance in the VarLen scenario. The reasons are clear: BERT-finetune as well as the related methods are not appropriate in VarLen scenario, especially when the target is longer than the input. The text generation models such as Transformer-s2s and GPT2-finetune suffer from the problem of hallucination \cite{DBLP:conf/acl/MaynezNBM20} and repetition, which are not steady on the problem of CGEC.

\subsection{Ablation Analysis}

\paragraph{Different Training Dataset}
Recall that we introduce several groups of training datasets in different scales as depicted in Table~\ref{tab:datasets}. It is also very interesting to investigate the performances on different-size datasets. Then we conduct training on those training datasets and report the results still on the SIGHAN2015 testset. The results are shown in Table~\ref{tbl:results_trainset}.
No matter what scale of the dataset is, TtT always obtains the best performance.

\paragraph{Impact of $\mathcal{L}_{\mathrm{dp}}$ and $\mathcal{L}_{\mathrm{crf}}$}

Table~\ref{tbl:results_loss} describes the performance of our model TtT and the variants without $\mathcal{L}_{\mathrm{dp}}$ (TtT w/o $\mathcal{L}_{\mathrm{dp}}$) and $\mathcal{L}_{\mathrm{crf}}$ (TtT w/o $\mathcal{L}_{\mathrm{crf}}$). We can conclude that the fusion of these two tasks, direct prediction and CRF-based dependency modeling, can indeed improve the performance.

\paragraph{Parameter Tuning for Focal Loss}

The focal loss penalty hyperparameter $\gamma$ is crucial for the loss function $\mathcal{L} = \mathcal{L}_{\mathrm{dp}} + \mathcal{L}_{\mathrm{crf}}$ and should be adjusted on the specific tasks~\cite{DBLP:journals/pami/LinGGHD20}. We conduct grid search for $\gamma \in (0, 0.1, 0.5, 1, 2, 5)$ and the corresponding results are provided in Table~\ref{tbl:results_g}. Finally, we select $\gamma = 0.5$ for TtT for the CGEC task.

\subsection{Computing Efficiency Analysis}
\begin{table}[!t]
\centering
\begin{tabular}{l|c|c}
\Xhline{2\arrayrulewidth}
 Model & Time (ms) & Speedup \\
 \hline
 Transformer-s2s & 815.40 &  1x   \\
 \hline
 GPT2-finetune & 552.82 & 1.47x    \\
 \hline
  TtT & 39.25 & 20.77x \\
 \hline
  BERT-finetune & 14.72 & 55.35x \\
 \hline
 \Xhline{2\arrayrulewidth}
\end{tabular}
\caption{Comparisons of the computing efficiency.}
\label{tab:efficiency}
\end{table}
Practically, CGEC is an essential and useful task and the techniques can be used in many real applications such as writing assistant, post-processing of ASR and OCR, search engine, etc. Therefore, the time cost efficiency of models is a key point which needs to be taken into account. Table~\ref{tab:efficiency} depicts the time cost per sample of our model TtT and some baseline approaches. The results demonstrate that TtT is a cost-effective method with superior prediction performance and low computing time complexity, and can be deployed online directly.

%\subsection{Case Analysis}

%\section{Ethical Considerations}
%Grammatical error correction is a useful task and can help people writing better. One possible issue may hide in the data augmentation procedure that some inappropriate sentences may be produced. Fortunately, it is can be easily fixed by some rule-based methods.

\section{Conclusion}
We propose a new framework named tail-to-tail non-autoregressive sequence prediction, which abbreviated as \textbf{TtT}, for the problem of CGEC. A BERT based sequence encoder is introduced to conduct bidirectional representation learning. In order to conduct substitution, deletion, insertion, and local paraphrasing simultaneously, a CRF layer is stacked on the up tail to conduct non-autoregressive sequence prediction by modeling the dependencies among neighbour tokens.
Low-rank decomposition and a truncated Viterbi algorithm are introduced to accelerate the computations.
Focal loss penalty strategy is adopted to alleviate the class imbalance problem considering that most of the tokens in a sentence are not changed.
Experimental results on standard datasets demonstrate the effectiveness of TtT in terms of sentence-level Accuracy, Precision, Recall, and F1-Measure on tasks of error Detection and Correction. TtT is of low computing complexity and can be deployed online directly.

In the future, we plan to introduce more lexical analysis knowledge such as word segmentation and fine-grained named entity recognition~\cite{DBLP:journals/corr/abs-2012-15639} to further improve the performance.

\bibliographystyle{acl_natbib}
\bibliography{acl2021}

%\appendix

\end{document}